# Diseño y desarrollo de aplicación móvil para la clasificación de flora nativa chilena utilizando redes neuronales convolucionales

*Design and development of a mobile application for the classification of Chilean native flora using convolutional neural networks.*

## Resumen

**Introducción:** Las aplicaciones móviles, a través de la visión artificial, son capaces de reconocer especies vegetales en tiempo real. Sin embargo, las actuales aplicaciones de reconocimiento de especies no consideran la gran variedad de especies endémicas y nativas de Chile, tendiendo a predecir erróneamente. Esta investigación presenta la construcción de un dataset de especies chilenas y el desarrollo de un modelo de clasificación optimizado e implementado en una aplicación móvil. **Método:** La construcción del dataset se realizó a través de la captura de fotografías de especies en terreno y selección de imágenes de datasets en línea. Se utilizaron redes neuronales convolucionales para desarrollar los modelos de predicción de imágenes. Se realizó un análisis de sensibilidad al entrenar las redes, validando con *k-fold cross validation* y realizando pruebas con distintos hiperparámetros, optimizadores, capas convolucionales y tasas de aprendizaje, para seleccionar los mejores modelos y luego ensamblarlos en un solo modelo de clasificación. **Resultados:** El dataset construido se conformó de 46 especies, incluyendo especies nativas, endémicas y exóticas de Chile, con 6120 imágenes de entrenamiento y 655 de prueba. Los mejores modelos se implementaron en una aplicación móvil, donde se obtuvo un porcentaje de acierto de aproximadamente 95% con respecto al conjunto de pruebas. **Conclusiones:** La aplicación desarrollada es capaz de clasificar especies correctamente con una probabilidad de acierto acorde con el estado del arte de la visión artificial y de mostrar información de la especie clasificada.

**Palabras clave:** Visión artificial; Redes neuronales convolucionales; Flora Chilena; Aplicaciones móviles.

## *Abstract*

*Introduction: Mobile apps, through artificial vision, are capable of recognizing vegetable species in real time. However, the existing species recognition apps do not take in consideration the wide variety of endemic and native (Chilean) species, which leads to wrong species predictions. This study introduces the development of a chilean species dataset and an optimized classification model implemented to a mobile app. **Method**: the data set was built by putting together pictures of several species captured on the field and by selecting some pictures available from other datasets available online. Convolutional neural networks were used in order to develop the images prediction models. The networks were trained by performing a sensitivity analysis, validating with k-fold cross validation and performing tests with different hyper-parameters, optimizers, convolutional layers, and learning rates in order to identify and choose the best models and then put them together in one classification model.*
***Results****: The final data set was compounded by 46 species, including native species, endemic and exotic from Chile, with 6120 training pictures and 655 testing pictures. The best models were implemented on a mobile app, obtaining a 95% correct prediction rate with respect to the set of tests.*
***Conclusion****: The app developed in this study is capable of classifying species with a high level of accuracy, depending on the state of the art of the artificial vision and it can also show relevant information related to the classified species.*

## Introducción

Los bosques de Chile cubren una superficie de 17,66 millones de hectáreas, lo que representa el 23,3 % de la superficie del territorio nacional. De esto, aproximadamente 14,41 millones de hectáreas equivalentes a un 19.04 % del territorio nacional son bosques nativos y 3,08 millones de hectáreas corresponden a plantaciones forestales (CONAF, 2017). La diversidad de especies vegetales presentes en Chile es muy grande, principalmente en la zona central debido a su ecosistema mediterráneo, lo que radica en un alto porcentaje de endemismo, en donde predomina el bosque esclerófilo.

A pesar de la gran biodiversidad de especies nativas presentes en Chile, los programas de arborización de espacios urbanos incluyen un mayor porcentaje de especies exóticas (Alvarado et al, 2013). Además, los textos escolares de ciencias naturales del país están constituidos por un 72% de especies de flora exótica en desmedro de la nativa y/o endémica. Esta incorporación de especies no nativas podría provocar un fenómeno llamado "homogeneización biocultural" y los niños son especialmente susceptibles a este, a causa de su poca interacción con la naturaleza y su actual y creciente conexión con medios digitales a la información global. La homogeneización biocultural "(...) *conlleva a un desconocimiento de la biodiversidad nativa con la que cohabitan por sobre especies carismáticas, que podría tener implicaciones en su apreciación o valoración por ella y, más aún, en los esfuerzos de conservación de esta.*" (Celis, 2016). Es difícil para las personas chilenas diferenciar especies nativas y exóticas, pudiendo confundir especies tan representativas como la Araucaria araucana con especies del género *Pinus* y *Eucalyptus* (Bravo et al., 2019). La falta de educación con respecto a las especies autóctonas radica en un mal reconocimiento de estas.

Las tecnologías de la información surgen como una alternativa viable para ayudar a las personas a reconocer especies en terreno.

## Visión artificial de flora

El problema de la clasificación automática de especies vegetales ya ha sido abordado hace algunos años. Existen diversos estudios y trabajos documentados que describen la construcción de algoritmos y sistemas que clasifican especies vegetales utilizando diferentes métodos (Cervantes et al., 2017). Los métodos utilizados analizan diferentes tipos de imágenes que contienen características de las especies, tal como las imágenes de hojas con fondo blanco, imágenes de hojas escaneadas, imágenes de hojas con fondo natural o imágenes que muestran un conjunto de hojas en el contexto natural de la especie. La precisión de los modelos varía entre un 80% y superando el 90%, dependiendo de la cantidad de especies clasificadas y los modelos construidos para su predicción. A pesar de que las metodologías y arquitecturas realizadas en estos estudios podrían ser aplicadas con especies nativas chilenas, son estudios que analizan conjuntos que no incluyen a las especies nativas o endémicas de Chile.

### Conjuntos de imágenes

Para poder desarrollar modelos de clasificación de especies vegetales es necesario contar con conjuntos de imágenes relativamente grandes para obtener resultados positivos de clasificación, en donde el modelo de predicción podrá reconocer sólo las especies que contenga este conjunto de imágenes.

Existen dataset de imágenes con sus clases correctamente etiquetadas disponibles en comunidades en línea dedicadas al aprendizaje automático (Kaggle, 2021), páginas oficiales de tecnologías ligadas al aprendizaje automático (Tensorflow, 2021) y de plataformas de crowdsourcing (ImageCLEF, 2017). Estos conjuntos de imágenes incluyen imágenes de algunas especies introducidas y exóticas en Chile y algunas pocas que corresponden a especies nativas de Chile, que también se encuentran en la distribución de otros países, sin embargo, no incluyen especies endémicas. Existen asociaciones de diferentes entidades que han realizado aplicaciones que permiten a los usuarios compartir imágenes de diversas especies del mundo a través de redes sociales especializadas para esto, en donde se pueden encontrar conjuntos de imágenes de especies nativas y endémicas de Chile, aunque estas imágenes están sujetas a los derecho de autor del usuario que las toma y no se presentan como un conjunto de imágenes, sino a imágenes individuales que, en algunas especies, no superan las 10 observaciones (iNaturalist, 2021).

### Aplicaciones de clasificación de flora

Existen actualmente aplicaciones para dispositivos móviles que son capaces de reconocer y diferenciar especies vegetales de diferentes partes del mundo con un buen porcentaje de clasificación, pero estas han sido desarrolladas en otros países enfocándose en flora de Europa, América del Norte y América Central (Bilyk et al, 2020.), lo cual podría radicar en una clasificación errónea de especies chilenas y desinformar a las personas al entregar resultados erróneos. Un ejemplo de aplicación líder en el mercado es PlantNet (Affouard et al, 2017), la cual clasifica más de 3700 especies de flora y que, en conjunto con otras entidades, realiza la competencia anual PlantClef (Goëau et al, 2017), la cual se enfoca en el desarrollo de nuevos modelos de aprendizaje supervisado para la clasificación de especies vegetales. Sin embargo, estas competencias y la aplicación se enfocan en especies mayoritariamente europeas con respecto a los dataset propuestos. Otro ejemplo es la aplicación PictureThis (Glority LLC, 2020), la cual es capaz de clasificar más de 10.000 especies vegetales distintas con una distribución variada. Esta aplicación tiene muy buenas opiniones, sin embargo, algunos usuarios han realizado sugerencias asegurando que la aplicación no es capaz de clasificar correctamente especies nativas chilenas (PictureThis, sin fecha). La principal ventaja de las aplicaciones de clasificación de flora sobre otras alternativas (e.g., páginas web) es que los usuarios pueden hacer la clasificación en terreno, en el momento en que encuentran la flora que desean clasificar. En esa situación, es más probable que el usuario tenga acceso a un teléfono móvil que a un computador, por lo que las aplicaciones móviles han sido la principal forma de implementación para los algoritmos de clasificación de flora.

**Propuesta de solución**

Dados los antecedentes expuestos, el presente trabajo propone el desarrollo de una aplicación móvil capaz de clasificar especies chilenas utilizando redes neuronales convolucionales. Para esto, proponemos construir un dataset de imágenes con especies que únicamente se encuentren dentro de la distribución geográfica chilena, y además propondremos el desarrollo de un modelo algorítmico de predicción optimizado para clasificar el dataset propuesto través de validación y pruebas para finalmente desplegar el modelo desarrollado en una aplicación móvil que sea accesible para toda la población.

# Marco Teórico

**Redes neuronales convolucionales**

Las redes neuronales convolucionales (CNN) son un tipo de red neuronal con aprendizaje supervisado y su aplicación más popular es la visión artificial, es decir, el reconocimiento de imágenes (Geron, 2019). Las CNN surgieron a partir de los estudios de la corteza visual (Hubel & Wiesel, 1962). Para el reconocimiento de imágenes la CNN contiene varias capas ocultas especializadas llamadas capas convoluciones. Estas capas permiten extraer características a través de una jerarquía en su distribución. Esto quiere decir que las primeras capas detectan patrones más simples y las capas más profundas reconocen patrones específicos y formas más complejas. El reconocimiento de patrones se realiza a través de filtros y kernels, los cuales mapean característica en base a los pixeles de las imágenes. Las redes neuronales convolucionales se componen principalmente por capas convolucionales que disminuyen la complejidad de las imágenes (tamaño) al profundizar la red y terminando esta con una capa completamente conectada, con una cantidad de neuronas igual a la cantidad de clases que se requiere clasificar. Las redes neuronales convolucionales pueden converger más rápidamente en su predicción y generalizar mejor los datos dependiendo del optimizador que se utilice, tasa de aprendizaje e hiperparámetros en general.

**ImageNet**

ImageNet es una gran base de datos visual que consta de millones de imágenes de diferentes clases (Deng et al., 2009). Se utiliza ampliamente en el área de investigación para el desarrollo de software de reconocimiento visual de objetos. La base de datos contiene más de 20.000 categorías. Anualmente, se realiza el concurso "ImageNet Large Scale Visual Recognition Challenge", donde se compite por realizar los mejores modelos clasificadores de imágenes con respecto a diversos retos. Es común en esta competencia el desafío de reconocer las 1.000 clases de ImageNet. Por esto, se considera que los resultados de los concursos de ImageNet corresponden al estado del arte del reconocimiento de imágenes.

Para medir la efectividad frente al dataset de 1.000 clases, se realiza una evaluación del porcentaje de aciertos frente a un conjunto de pruebas. Las medidas son Top-1 accuracy, el cual corresponde al porcentaje de aciertos y Top-5 accuracy que corresponde al porcentaje de aciertos tomando en cuenta como válido si la clase original de la imagen se encuentra dentro de los 5 mejores accuracy dado el resultado del modelo de clasificación. Actualmente, el estado del arte con respecto al conjunto ImageNet es de un 90.2% de Top-1 accuracy (Pham et crossal., 2020).

**MobileNet**

MobileNet es una arquitectura de red neuronal convolucional, la cual ha sido desarrollada con el objetivo de implementarse en dispositivos embebidos y móviles. Utilizan solamente dos hiperparámetros, por lo que son relativamente simples de utilizar y al mismo tiempo logran buenos *accuracy* (Howard et al., 2017). Actualmente existen dos arquitecturas principales de MobileNet: MobileNetV1 (Howard et al., 2017) y MobileNetV2 (Sandler et al., 2018).

**PlantClef**

ImageCLEF se lanzó en 2003 como parte del *Cross Language Evaluation Forum* (CLEF) y tiene como objetivo principal apoyar el avance del campo del análisis, indexación, clasificación y recuperación de medios visuales. Anualmente se realizan competencias donde los participantes realizan modelos de predicción de imágenes basados en diversas áreas y dataset. PlantClef es parte de ImageClef y es una subdivisión que se enfoca en la clasificación de especies vegetales de diversos lugares del mundo. La versión del año 2017 obtuvo excelentes resultados con respecto a la clasificación de especies del oeste de Europa y Norte América con un dataset de 1.1 millones de imágenes para 10.000 clases (PlantClef, 2017) .

## Metodología

**Construcción de dataset**

La obtención de imágenes se realizó utilizando dos métodos distintos:

a) Investigación a través de diversas plataformas web relacionadas al aprendizaje automático y repositorios de imágenes de plantas.
b) Se realizó una identificación de árboles nativos, endémicos y exóticos utilizando guías de identificación de árboles y vegetación, además con el apoyo de un profesional del área. Al identificar los árboles, se realizaron capturas fotográficas en diferentes ángulos de hojas individuales, flores, frutos y una combinación de estos.

Se utiliza un mínimo de 100 imágenes por clases. La identificación y selección de las especies se realizó utilizando una guía de árboles nativos de Chile (Alvarado et al., 2013), guía de árboles urbanos utilizados para programas de arborización (García & Ormazabal, 2008) y una investigación de especies introducidas en el país (Rodrigues, G. & Rodriguez, R., 1981).

**Extracción de imágenes de datasets públicos**

La extracción de imágenes se realizó utilizando el dataset propuesto para la competencia

PlantCLEF 2017 con más de 10.000 especies, extrayendo especies relacionadas a la distribución chilenas. Ya que las imágenes disponibles presentan distintas dimensiones, se desarrolló un algoritmo en Python (Muñoz, 2021), utilizando la librería OpenCv para recortar las imágenes y redimensionarlas.

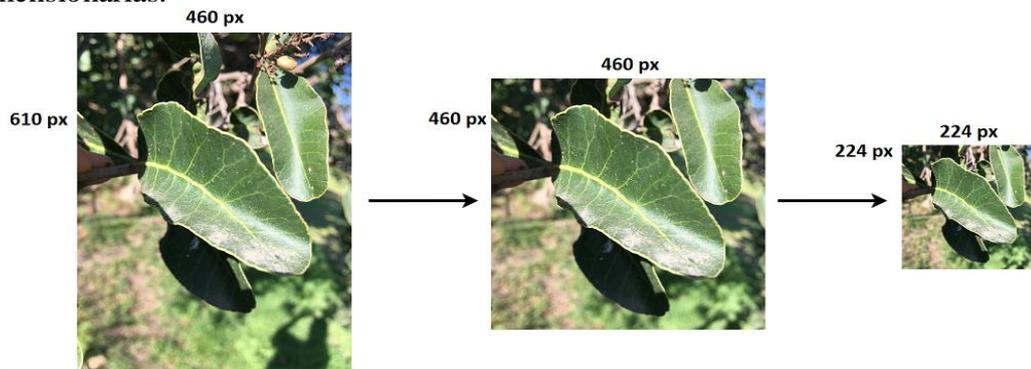

**Figura 1.** Recorte y redimensionamiento de imágenes de PlantCLEF 2017

Según la proporción entre alto y ancho, se recortan los extremos horizontales o verticales como se muestra en figura 1, para no deformar la imagen al redimensionar y se redimensiona a 224 x 224 píxeles, formato de pixeles compatible con arquitecturas MobileNet.

**Captura de imágenes en terreno**

La captura de imágenes se realizó en sectores urbanizados y rurales de la zona central de Chile, específicamente en la provincia de Santiago:

a)   Cerro Chena, San Bernardo
b)   Cuesta Barriga, Talagante
c)   Cerro Tantehue, Melipilla
d)   Zonas urbanas, Maipú

Para la captura de fotos se usó una cámara de smartphone con una resolución de 3024 x 3024 píxeles. La cantidad de pixeles de ancho y alto es directamente proporcional, ya que las imágenes utilizadas se consideran en bruto, y debieron ser redimensionadas sin recortar a 224 x 224 píxeles antes de construir el dataset.

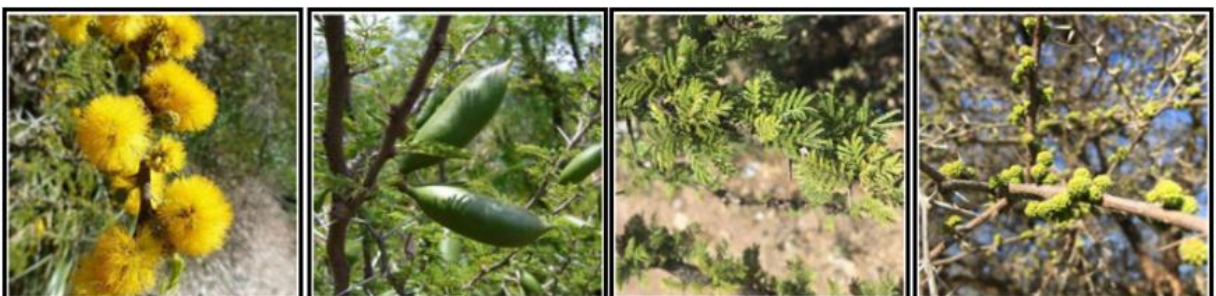

**Figura 2.** Órganos capturados de especie Vachellia Caven. De izquierda a derecha: flores, semillas, hojas y brotes florales.

Para aumentar la variedad características observables de una planta y así disminuir el *overfitting* de nuestros modelos (i.e., para que no se enfoque en una característica particular de la planta), se capturan fotos de cada especie con distintos contextos de fondo y a distintos órganos del espécimen, tal como se muestra en Figura 2.

**Validación modelo óptimo**

La metodología de desarrollo de las redes neuronales convolucionales correspondientes a los modelos de predicción del sistema corresponde a una metodología experimental basada en iteraciones. Cada iteración corresponde a un conjunto de tareas para producir un modelo óptimo basado en arquitecturas MobileNet.

Para la construcción de las redes neuronales convolucionales se llevó a cabo un proceso experimental que culmina en la validación de las arquitecturas óptimas para un modelo en base a su porcentaje de aciertos o *accuracy* de clasificación para diferentes conjuntos de validación y, finalmente, un conjunto de test.

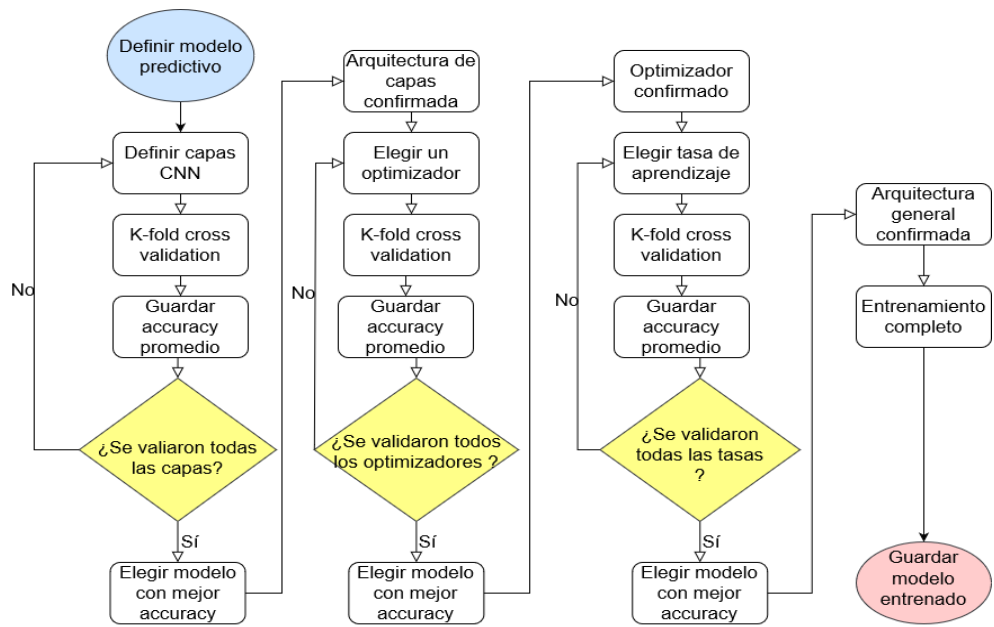

**Figura 3.** Metodología para validar arquitectura óptima

La secuencia de acciones para obtener la configuración de CNN que maximice la exactitud de clasificación se muestra en el diagrama de flujo en figura 3. Los pasos a seguir son los siguientes:

a)  Se realiza un preprocesamiento de las imágenes en donde estas se separan por clase, además de redimensionar y generar aleatoriamente copias de las imágenes con características distintas como volteo y zoom para aumentar la cantidad de imágenes del conjunto de datos.
b)  Se construyen diferentes arquitecturas de redes neuronales con respecto a las capas que la componen. Cada arquitectura se entrena y valida utilizando *k-fold cross validation* y se selecciona la arquitectura con mejor accuracy de validación.

c)   En base a la arquitectura de capas escogida, se realizan pruebas utilizando *k-fold cross validation* para diferentes optimizadores y se selecciona el mejor optimizador para la arquitectura CNN.

d)   En base a la arquitectura de capas y optimizador escogido, se realizan pruebas utilizando *k-fold cross validation* de distintas tasas de aprendizaje. Finalmente es seleccionada la mejor tasa de aprendizaje.

e)   La arquitectura general de la CNN se valida y se entrena con respecto a la totalidad del conjunto de entrenamiento, para ser guardada y probada posteriormente con respecto a un conjunto de test.

**Ensamble modelos MobileNet**

Como parte de los modelos realizados, se construye un ensamble de redes neuronales utilizando los modelos individuales para crear un único modelo.

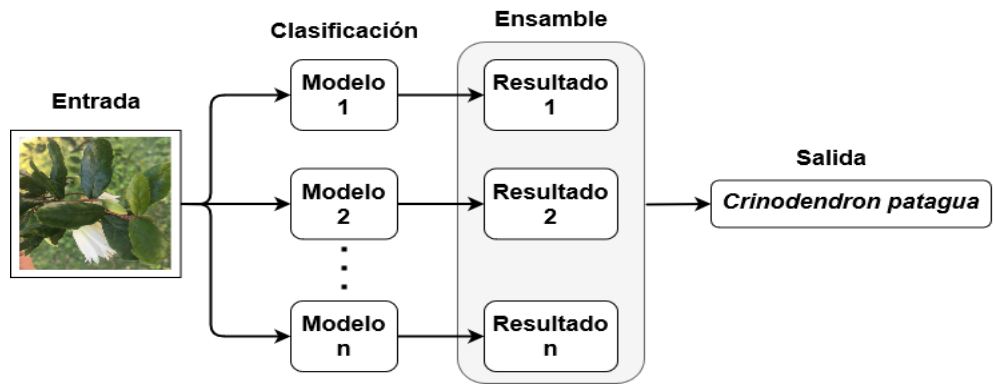

**Figura 4.** Funcionamiento ensamble

Un modelo individual, al realizar una predicción, produce una lista de probabilidades, donde la cantidad de probabilidades individuales corresponde a la cantidad de clases con las cuales se ha entrenado el modelo y cada clase se relaciona con una probabilidad en una posición específica dentro de la lista. El método de ensamble mostrado en la Figura 4, promedia la probabilidad de cada clase individual entre los modelos utilizados, donde cada clase dentro de la lista de predicción se encuentra en la misma posición para cada modelo. Se calcula la probabilidad de individual de una clase en específico a través de:

$$\%c_i e_j = \frac{\%c_1 m_i + \%c_2 m_i}{m} \quad (1)$$

Donde $\%c_i e_j$ representa la probabilidad de que una clase $c_i$ en la posición $i$ dado el modelo del ensamble $e_j$ corresponda a la imagen de entrada dados los modelos $m_1$ y $m_2$, donde m corresponde a la totalidad de modelos de entrada.

De esta forma, se produce una única lista de probabilidades por clase, de donde se selecciona la máxima probabilidad la cual se relaciona según su posición con una clase en específico. La máxima probabilidad se relaciona según su posición relacionada con una clase y corresponde a la clase clasificada.

$$\max( [ \%_{c_i}, \%_{c_{i+1}}, \%_{c_{i+2}}, \ldots, \%_{c_{n-1}}, \%_{c_n} ]) \qquad (2)$$

## Desarrollo y resultados

### Conjunto entrenamiento y pruebas

Las especies extraídas del conjunto de imágenes PlantClef 2017 fueron 11, las cuales corresponden a especies exóticas, pero que se encuentran en la distribución chilena según. Con respecto a la captura de imágenes en terreno se logró capturar imágenes de 35 especies exóticas, nativas y endémicas de Chile. Finalmente, el dataset se compone de 46 especies con 6775 imágenes distintas en total, donde el conjunto de pruebas se compone de un 10% aproximado del total de imágenes.

Tabla 1
**Imágenes por conjunto de entrenamiento y pruebas**

| Conjunto | Porcentaje (%) | N° imágenes |
|---|---|---|
| Entrenamiento | 90.33 | 655 |
| Pruebas | 9.67 | 6120 |
| Dataset | 100.00 | 6775 |

El etiquetado de imágenes se realizó a través de documentos en formato CSV, relacionando una clase (especie) a cada nombre de imagen. El dataset se encuentra disponible para su uso público (Muñoz, 2020).

### Entrenamiento y validación

Se usa la metodología mostrada en figura 4 para encontrar el mejor modelo dadas las arquitecturas Mobilenet propuestas. Se utilizó *k-fold cross validation* para validar los modelos separando el conjunto de entrenamiento en cinco particiones. Así mismo, la medida de exactitud utilizada es *Top-1 accuracy*. Las pruebas consideran los siguientes casos:

a) Capa densa extra: se añade a la arquitectura de la red una capa densa o completamente conectada justo antes de la capa final de clasificación Softmax. Dicha capa será de 256 o 512 neuronas.
b) Optimizadores: se utilizan los optimizadores Adam, Adamax, Adagrd y SGD
c) Tasa de aprendizaje: se asignan distintos valores de tasa de aprendizaje en un rango de 0.005 a 0.01.

Los experimentos se realizaron utilizando 30 épocas y un *batch size* de 16 para el entrenamiento y la validación.

Validación capas densas

Los modelos tienden a converger cerca de la época 20. Como se observa en el Gráfico 1, MobileNet obtiene mejores resultados para todas las configuraciones de capas, en donde el mayor Top-1 accuracy es de 89,97% al agregar una capa densa de 512 antes de la capa Softmax final. La configuración original de MobileNetV2 obtiene un 89,48%.

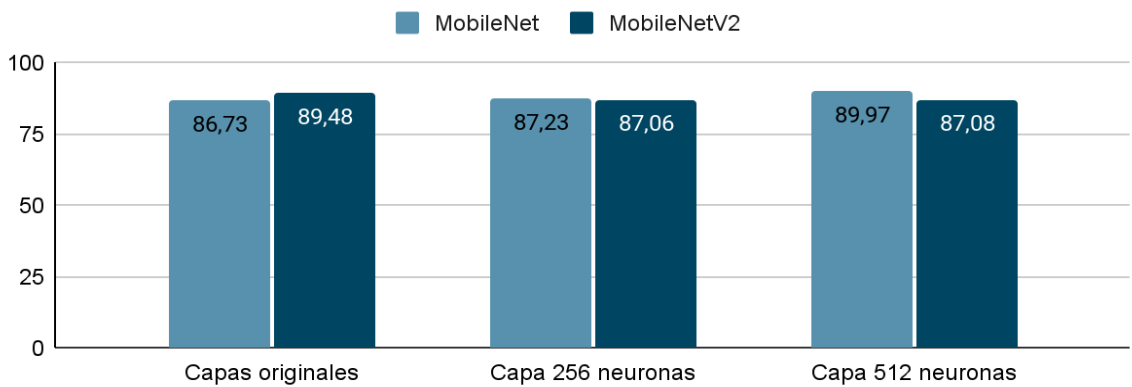

**Gráfico 1.** MobileNet v/s MobileNetV2: resultados validación cruzada capas densas

Validación optimizadores

Se mantuvo la arquitectura original de MobileNet para las pruebas, debido a los resultados obtenidos en Gráfico 1. El Top-1 accuracy utilizando el optimizador Adam corresponde al optimizador con menor accuracy para ambas arquitecturas MobileNet y converge de forma irregular. Para el caso de Adamax, SGD y Adagrad, los accuracy se encuentran cercanos al 90%. Los mayores accuracy para ambas arquitecturas MobileNet corresponden al optimizador Adagrad.

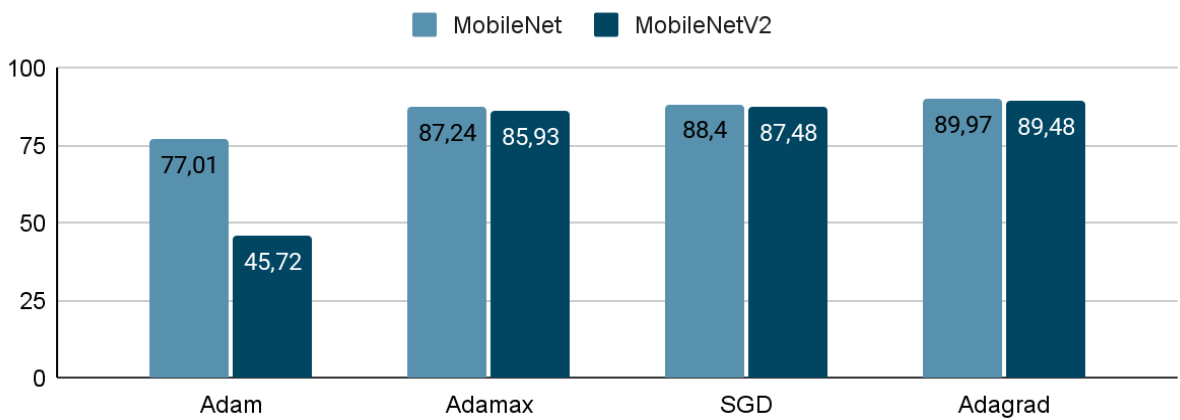

**Gráfico 2.** MobileNet v/s MobileNetV2: resultados validación cruzada optimizadores

Validación tasas de aprendizaje

Se realizan las pruebas utilizando el optimizador Adagrad. Los resultados mostrados en el Gráfico 3 indican que, para MobileNetV2, la variación del accuracy de validación no es muy grande. El mayor accuracy de validación obtenido para MobileNetV2 es de 89,59% utilizando una tasa de aprendizaje de 0.005. MobileNet obtiene un mayor accuracy de 90.13% de validación utilizando una tasa de aprendizaje de 0.005.

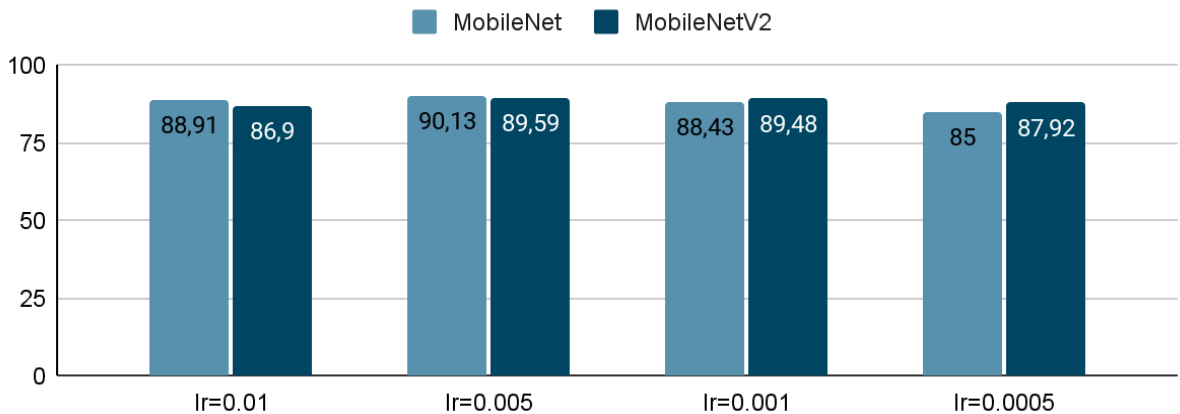

**Gráfico 3.** MobileNet v/s MobileNetV2: resultados validación cruzada tasa de aprendizaje

**Arquitecturas óptimas**

En base a los experimentos realizados, se define la configuración óptima de capas, optimizador y tasa de aprendizaje para MobileNet y MobileNetV2.

Tabla 2
**Configuración óptima de arquitecturas Mobilenet para dataset propuesto**

| Tipo | Nombre | Capa densa | Optimizador | Tasa aprendizaje |
| --- | --- | --- | --- | --- |
| MobileNet | MNetV1D512 | 512 | Adagrad | 0.005 |
| MobileNetV2 | MNetV2 | Original | Adagrad | 0.005 |

Con las configuraciones definidas en Tabla 2, se realiza un entrenamiento completo de 40 épocas con respecto al conjunto de entrenamiento. Una vez entrenado cada modelo, se guardan y se construye el ensamble de redes MobileNet utilizando las votaciones ponderadas indicadas en (1) y (2) considerando ambos modelos base. El modelo de ensamble fue nombrado como "MNetEnsamble".

**Resultados conjunto de pruebas**

Las pruebas se realizaron con respecto al conjunto de pruebas conformado por 655 imágenes. Se convirtieron los modelos Keras a formato Tensorflow Lite utilizando el método de cuantificación posterior al entrenamiento (Tensorflow, 2021).

Se observa en Tabla 3 que el tamaño de los modelos disminuye aproximadamente a la mitad al convertirlos a formato Tensorflow Lite. El modelo MNetEnsamble está conformado por los modelos MNetV1D512 y MNetV2, por lo que su peso corresponde a la suma de los pesos de ambos modelos.

Tabla 3
**Peso en Megabytes de modelos**

| Modelo | Tamaño modelo Keras (Mb) | Tamaño modelo TF Lite (Mb) |
| --- | --- | --- |
| MNetV1D512 | 349.0 | 174.0 |
| MNetV2 | 62.7 | 30.9 |
| MNetEnsamble | 411.7 | 204.9 |

Como se observa en Gráfico 4, los tres modelos finalistas obtuvieron un *Top-1 accuracy* de la clasificación del conjunto de pruebas sobre el 90%. El mejor *accuracy* obtenido corresponde al modelo MNetEnsamble, con un 94.8% correspondiente al modelo Keras y un 95.1% para el modelo Tensorflow Lite ya testeado en la aplicación. El *accuracy* no varía mayormente al convertir los modelos de Keras a TensorFlow Lite y la tendencia es la misma, en donde los *accuracy* de pruebas en orden de menor a mayor corresponden a MNetV2, MNetV1D512 y MNetEnsamble respectivamente.

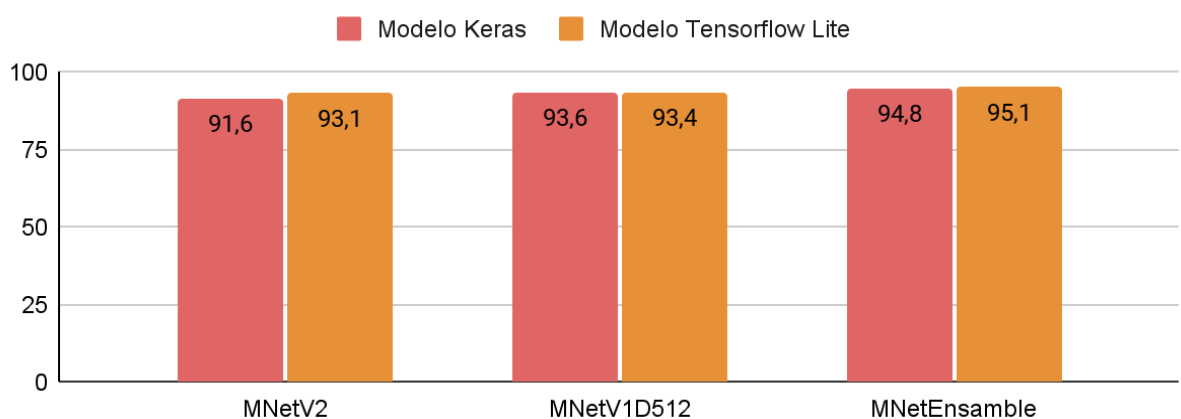

**Gráfico 4.** Modelo Keras v/s Tensorflow: resultados conjunto de pruebas

**Especificación de requerimientos**

Se realizó un análisis de requerimientos en base a las principales aplicaciones de reconocimiento de especies vegetales, generando casos de uso que definen las funcionalidades

de la aplicación. A continuación, se muestra un diagrama de casos de uso en notación UML (OMG, 2017) que resume los casos de uso contenidos en esta aplicación:

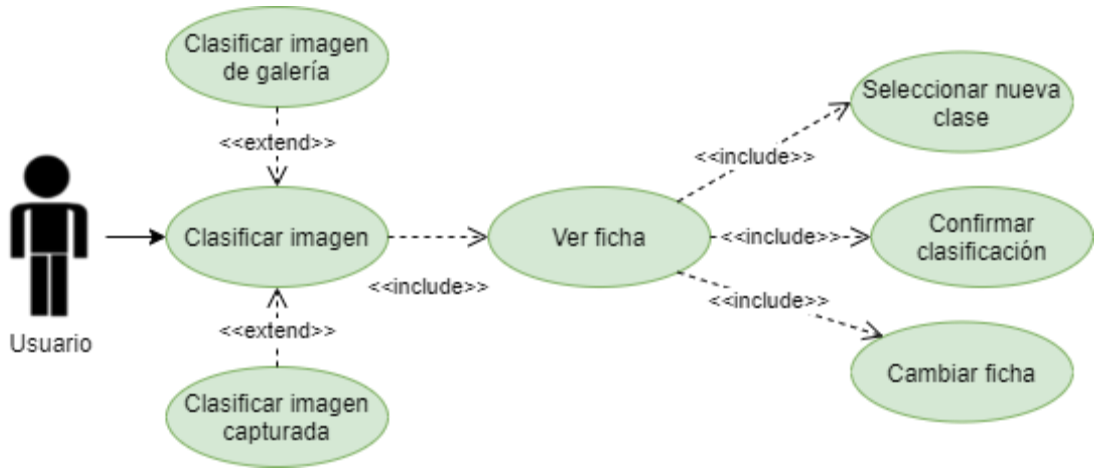

**Figura 5.** Casos de uso

El sistema se desarrolló como un prototipo con funcionalidades clave, enfocadas en la clasificación de la especie y entrega de información, como se observa en la Figura 5. El código fuente se encuentra disponible de forma pública (Muñoz, 2020).

**Implementación y funcionalidades de la aplicación**

La aplicación desarrollada permite la selección de imágenes en el almacenamiento del dispositivo o captura de fotos en tiempo real, tal como se observa en la Figura 6. Cuando el usuario selecciona una fotografía o la captura, la aplicación realiza la clasificación de la imagen basándose en el modelo MNetEnsamble implementado.

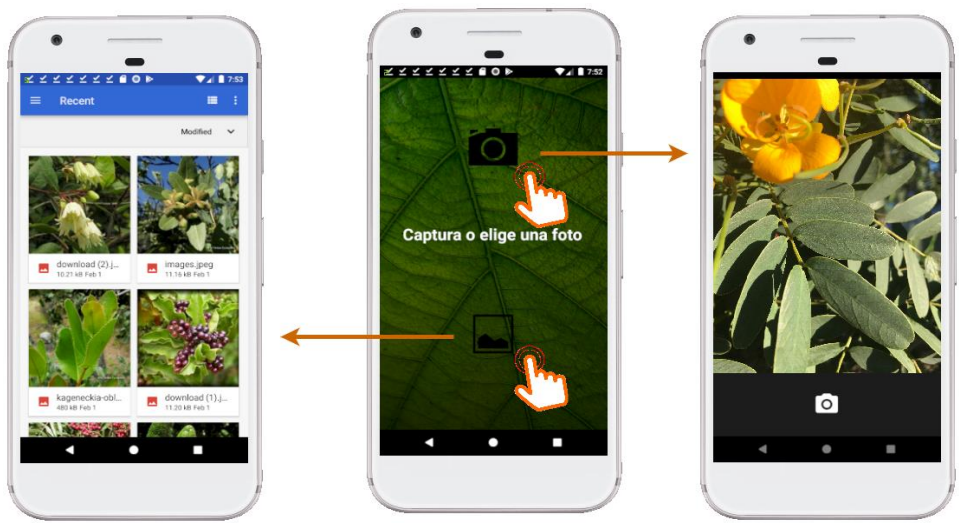

**Figura 6.** Pantalla de inicio

Al realizar la clasificación, la aplicación muestra en pantalla los datos de la clasificación realizada junto con una imagen de la especie clasificada y una miniatura de la captura realizada con el objetivo de poder compararlas. Los datos mostrados en pantalla son consultados a través de una base de datos embebida en la aplicación que contiene los principales datos de las especies como nombres comunes, nombre científico/botánico, probabilidad de acierto, tipo de especie (endémica, nativo o exótica), estado de conservación, distribución geográfica y descripción.

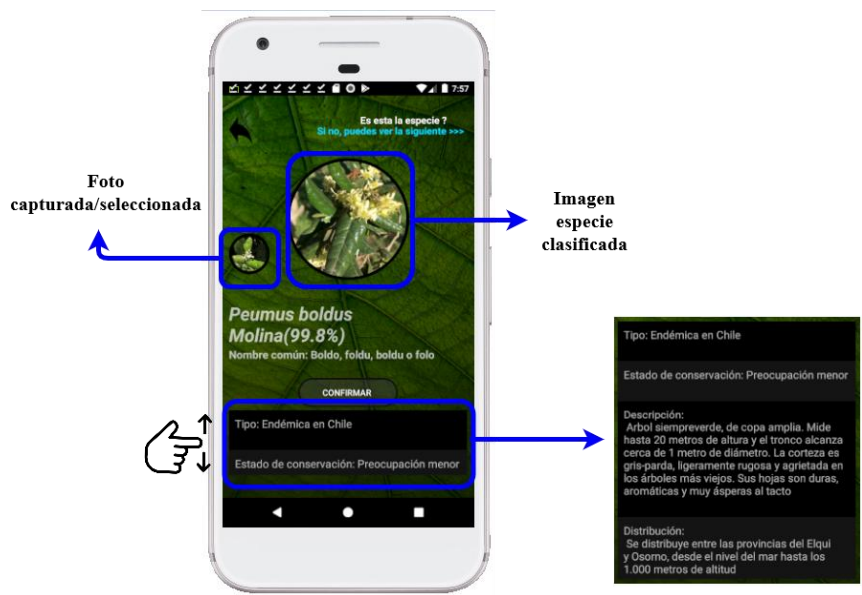

**Figura 7.** Pantalla con especie clasificada

La pantalla mostrada en Figura 7 permite ver más de una clasificación en caso de un posible error de clasificación, mostrando más opciones en orden decreciente con respecto a la probabilidad de acierto de la especie. El usuario tiene la opción de confirmar la especie o de seleccionar a través de un combobox de una lista de especies posibles, para poder retroalimentar los resultados.

**Discusión de resultados**

Los resultados obtenidos con respecto al conjunto de pruebas son bastante buenos (95% *accuracy* aprox) y están al nivel del estado del arte de la visión artificial (Hieu et al., 2020.) y superando el Top-1 accuracy de las arquitecturas MobileNet. Sin embargo, la cantidad de clases que se clasifican basándose en el estado del arte de ImageNet son 10.000, una gran diferencia con respecto a las 46 clases de los modelos desarrollados. La cantidad de especies vegetales en Chile es de 2.630 especies endémicas, 2.452 nativas y 657 exóticas. Por lo tanto, las clases totales de la aplicación y que debería poder clasificar son aproximadamente 5.000.

El método de ensamble de arquitecturas MobileNet obtuvo resultados cercanos al estado del arte y logró maximizar el *accuracy* de los modelos individuales. Esta mejora de *accuracy* ensamblando los resultados de redes individuales se puede ver reflejada también en otros trabajos, por ejemplo, el ganador de la competencia PlantCLEF 2017, el cual obtuvo un accuracy de 88.5% utilizando métodos de ensamble, específicamente un Bagging modificado con 12 modelos distintos en base a 3 arquitecturas conocidas (Lasseck, 2017). Este trabajo realizado es una buena base para mejorar los modelos desarrollados, en donde se podría incluir al ensamble más modelos de distintas arquitecturas livianas, como por ejemplo, arquitecturas de tipo EfficientNet. El ensamble desarrollado en este trabajo es capaz de integrar más de dos modelos, por lo que sería beneficioso entrenar más modelos e implementarlos.

## Conclusiones y trabajo futuro

La construcción del dataset se logró con un total de 6775 imágenes y 46 especies, en donde 5 son nativas, 13 endémicas y 28 exóticas, todas dentro de la distribución chilena. Las mejores configuraciones para el dataset propuesto corresponden a las mostradas en la tabla 2, las cuales aumentan la generalización de los datos.

Con respecto al método de ensamble, se comprueba que logra maximizar el accuracy, obteniendo mejores resultados que los modelos individuales de tipo MobileNet.

Los modelos desarrollados, al ser convertidos en formato Tensorflow Lite disminuyeron su tamaño, pero no tuvieron un gran cambio con respecto a su accuracy frente al conjunto de validación. Esto podría deberse a que se utilizó cuantificación como método para limitar los bits de precisión de los parámetros del modelo, lo cual no necesariamente podría conllevar a una disminución del accuracy, ya que la precisión varía dependiendo del modelo al realizar la conversión y *"(...) muchos pequeños pasos en la dirección correcta dan como resultado mejores predicciones con datos de prueba. Mediante la cuantificación, es posible obtener una precisión mejorada debido a la menor sensibilidad de los pesos."* (Cawin, 2021).

Finalmente, se concluye que el objetivo principal del trabajo realizado ha sido logrado, permitiendo así clasificar especies a través de fotos en tiempo real y mostrando al usuario sus datos más importantes y definiendo si son nativos, endémicos o exóticos.

Para el trabajo futuro se proyecta el desarrollo de un servicio web que permita almacenar las imágenes capturadas por los usuarios y los datos de clasificación realizados por la aplicación. De esta manera se pretende aumentar la cantidad de imágenes y clases (i.e., especies) disponibles en los modelos, reentrenando con los nuevos datos. También se considera la implementación de una interfaz de usuario para administrar los servicios y verificar las imágenes subidas y su clasificación.

También se desarrollará un método de captura de imágenes más eficiente, capturando imágenes en terreno con cámaras de alta resolución y recortando a modo de grilla con un script la imagen, obteniendo así más de una imagen por foto, agilizando el proceso de inclusión de especies poco comunes.

# REFERENCIAS